\documentclass{svjour3-custom}

\usepackage{graphics}
\usepackage{epsfig}
\usepackage{amsmath}

\numberwithin{equation}{section}

\journalname{\emph{Genetic Programming and Evolvable Machines},}

\begin{document}

\title{Expert-Driven Genetic Algorithms for Simulating Evaluation Functions\footnote{A preliminary version of this paper appeared in  \emph{Proceedings of the 2008 Genetic and Evolutionary Computation Conference} \cite{david08a} and received the Best Paper Award in the conference's Real-World Applications track.}}

\titlerunning{Expert-Driven GA for Simulating Evaluation Functions}

\author{Eli (Omid) David \and Moshe Koppel \and \\Nathan S. Netanyahu}

\authorrunning{E.O. David, M. Koppel, N.S. Netanyahu}

\institute{E.O. David\at
Department of Computer Science, Bar-Ilan University, Ramat-Gan 52900, Israel\\
\email{mail@elidavid.com}, Website: www.elidavid.com
\and
M. Koppel \at
Department of Computer Science, Bar-Ilan University, Ramat-Gan 52900, Israel\\
\email{koppel@cs.biu.ac.il}
\and
N.S. Netanyahu \at
Department of Computer Science, Bar-Ilan University, Ramat-Gan 52900, Israel, and Center for Automation Research, University of Maryland, College Park, MD 20742\\
\email{nathan@\{cs.biu.ac.il, cfar.umd.edu\}}
}

\date{}

\maketitle

\abstract{In this paper we demonstrate how genetic algorithms can be used to reverse engineer an evaluation function's parameters for computer chess. Our results show that using an appropriate expert (or mentor), we can evolve a program that is on par with top tournament-playing chess programs, outperforming a two-time World Computer Chess Champion. This performance gain is achieved by evolving a program that mimics the behavior of a superior expert. The resulting evaluation function of the evolved program consists of a much smaller number of parameters than the expert's. The extended experimental results provided in this paper include a report of our successful participation in the 2008 World Computer Chess Championship. In principle, our expert-driven approach could be used in a wide range of problems for which appropriate experts are available.\keywords{Computer chess, Fitness evaluation, Games, Genetic algorithms, Parameter tuning}}

\section{Introduction}

Since the dawn of modern computer science, game playing has posed a formidable challenge in the field of Artificial Intelligence. Many founding figures of computer science and AI (including Alan Turing, Claude Shannon, Konrad Zuse, Arthur Samuel, John McCarthy, Ken Thompson, Herbert Simon, and others) developed game-playing programs and used games in AI research.

The ongoing key role played by and the impact of computer games on AI should not be underestimated. If nothing else, computer games have served as an important testbed for spawning various innovative AI techniques in domains and applications such as search, automated theorem proving, planning, and learning. In addition, the annual World Computer Chess Championship (WCCC) is arguably the longest ongoing performance evaluation of programs in computer science, which has inspired other well-known competitions in robotics, planning, and natural language understanding.

Computer chess, while being one of the most researched fields within AI, has not lent itself to the successful application of conventional learning methods, due to its enormous complexity. Hence, top chess programs still resort to manual tuning of the parameters of their evaluation function. The latter assigns a score to a given chess position and is thus the most critical component of any chess program.

In this paper, we introduce a novel \emph{expert-driven} approach for automatically evolving the parameters of a chess program's evaluation function through the use of genetic algorithms (GA). The results show that our expert-driven approach for the application of GA efficiently evolves these parameters from randomly initialized values to highly tuned ones, yielding a program that outperforms its original version by a wide margin. Such performance was achieved for an evolved program whose evaluation function is considerably smaller than the expert's, in terms of its number of parameters.

In this paper, we extend results provided in earlier work \cite{david08a} to provide an in-depth assessment of the performance gain due to evolution. These experiments consist primarily of a longer series of matches played between the evolved organism and the expert, to compare the performance with higher statistical confidence. (A detailed  quantitative derivation is provided to that effect in Appendix B.) Additionally, we compare the performance of the initial random organisms to that of the expert and the evolved organism, and the performance of the evolved organism against several top commercial chess programs, including its relative tactical strength with respect to a suite of tactical test positions. Finally, we provide a detailed account of our participation in the 2008 World Computer Chess Championship. Running on an average single processor laptop against nine of the strongest programs in the world (eight of which ran on fast multicore machines ranging from 4 to 40 cores), our genetically evolved program reached second place in the World Computer Speed Chess Championship, and sixth place in the World Computer Chess Championship, thereby further establishing the practical merit of our approach. 

The rest of the paper is organized as follows. In Section 2 we review past attempts at applying evolutionary techniques in computer chess. We also compare alternative learning methods to evolutionary methods, and argue why the latter are more appropriate for the task in question. Section 3 presents our expert-driven approach, including a detailed description of the chess programs used and the framework of the GA as applied to the problem. Section 4 provides our experimental results, and Section 5 contains concluding remarks and suggestions for future research.

\section{Learning in Computer Chess}

The enormously complex game of chess, referred to as ``the touchstone of the intellect'' by Goethe, has always been one of the main battlegrounds of man versus machine. John McCarthy refers to chess as the ``Drosophila of AI'' \cite{mccarthy90}. Chess-playing programs have come a long way over the past several decades. While the first chess programs could not pose a challenge to even a novice player, the current advanced chess programs are on par with the strongest human chess players, as the recent man vs.~machine matches clearly indicate. This improvement is largely a result of deep searches that are possible nowadays, thanks to both hardware speed and improved search techniques. While the search depth of early chess programs was limited to only a few plies, nowadays tournament-playing programs easily search more than a dozen plies in middlegame, and tens of plies in late endgame.

Despite their groundbreaking achievements, a glaring deficiency of today's top chess programs is their severe lack of a learning capability (except in most negligible ways, e.g., ``learning'' not to play an opening that resulted in a loss, etc.). In other words, despite their seemingly intelligent behavior, those top chess programs are mere brute-force (albeit efficient) searchers that lack true underlying intelligence.

\subsection{Conventional vs. Evolutionary Learning in Computer Chess}

During more than fifty years of research in the area of computer games, many learning methods such as reinforcement learning \cite{sutton98} have been employed in simpler games. Temporal difference learning has been successfully applied in backgammon and checkers \cite{schaeffer01,tesauro92}. Although temporal difference learning has also been applied to chess \cite{baxter00}, the results showed that after three days of learning, the playing strength of the program was merely 2150 Elo (see Appendix B for a description of the Elo rating system), which is a very low rating for a chess program. In a more recent paper, Block \emph{et al.}~\cite{block08} reported on their experiments applying reinforcement learning to chess. Their results show that after the learning and improvement phase, their program achieves a playing strength of only 2016 Elo, which is amongst the lowest ratings for any chess program. Wiering \cite{wiering95} provided formal arguments for the failure of these methods in more complicated games such as chess.

The issue of learning in computer chess is basically an optimization problem. Each program plays by conducting a search, where the root of the search tree is the current position, and the leaf nodes (at some predefined depth of the tree) are evaluated by some static evaluation function. In other words, sophisticated as the search algorithms may be, most of the knowledge of the program lies in its evaluation function. Even though automatic tuning methods, that are based mostly on reinforcement learning, have been successfully applied to simpler games such as checkers, they have had almost no impact on state-of-the-art chess engines. Currently, all top tournament-playing chess programs use hand-tuned evaluation functions, since conventional learning methods cannot cope with the enormous complexity of the problem. This is underscored by the following points:

(1) The space to be searched is huge. It is estimated that there are up to $10^{46}$ possible positions that can arise in chess \cite{chinchalkar96}. As a result, any method based on exhaustive search of the problem space is infeasible.

(2) The search space is not smooth and unimodal. The evaluation function's parameters of any top chess program are highly co-dependent. For example, in many cases increasing the values of three parameters will result in a worse performance, but if a fourth parameter is also increased, then an improved overall performance would be obtained. Since the search space is not unimodal, i.e., it does not consist of a single smooth ``hill'', any gradient-ascent algorithm such as hill climbing will perform poorly. Genetic algorithms, on the other hand, are known to perform well in large search spaces which are not unimodal.

(3) The problem is not well understood. As will be discussed in detail in the next section, even though all top programs are hand-tuned by their programmers, finding the best value for each parameter is based mostly on educated guessing and intuition. (The fact that all top programs continue to operate in this manner attests to the lack of practical alternatives.) Had the problem been well understood, a domain-specific heuristic would have outperformed a general-purpose method such as GA.

(4) We do not require a global optimum to be found. Our goal in tuning an evaluation function is to adjust its parameters so that the overall performance of the program is enhanced. In fact, a unique global optimum does not exist for this tuning problem.

In view of the above points it seems appropriate to employ GA for automatic tuning of the parameters of an evaluation function. Indeed, at first glance this appears like an optimization task, well suited for GA. The many parameters of the evaluation function (bonuses and penalties for each property of the position) can be encoded as a bit-string. We can randomly initialize many such ``chromosomes'', each representing one evaluation function. Thereafter, one needs to evolve the population until highly tuned ``fit'' evaluation functions emerge.

However, there is one major obstacle that hinders the above application of GA, namely the fitness function. Given a set of parameters of an evaluation (encoded as a chromosome), how should the fitness value be calculated? For many years, it seemed that the solution was to let the individuals, at each generation, play against each other a series of games, and subsequently, record the score of each individual as its fitness value. (Each ``individual'' is a chess program with an appropriate evaluation function.)

The main drawback of this approach is the unacceptably large amount of time needed to evolve each generation. As a result, severe limitations were imposed on the length of the games played after each generation, and also on the size of the population involved. With a population size of 100, a limitation of 1 minute per game for each side, and assuming that each individual plays at least 10 games, it would take 2000 minutes for each generation to evolve. Specifically, reaching the 50th generation would take no less than 70 days.

In Section 3 we present our expert-driven approach for using GA in state-of-the-art chess programs. Before that, we briefly review previous work in applying evolutionary methods in computer chess.

\subsection{Previous Evolutionary Methods Applied to Chess}

Despite the abovementioned problems, there have been some successful applications of evolutionary techniques in computer chess, subject to some restrictions. Genetic programming was successfully employed by Hauptman and Sipper \cite{hauptman05,hauptman07} for evolving programs that can solve Mate-in-N problems and play chess endgames.

Kendall and Whitwell \cite{kendall01} used evolutionary algorithms for tuning the parameters of an evaluation function. Their approach had limited success, due to the very large number of games required (as previously discussed), and the small number of parameters used in their evaluation function. Their evolved program managed to compete with strong programs only if their search depth (lookahead) was severely limited.

Similarly, Aksenov \cite{aksenov04} used genetic algorithms for evolving the parameters of an evaluation function, using games between the organisms for determining their fitness. Again, since this method required a very large amount of games, the method evolved only a few parameters of the evaluation function with limited success. Tunstall-Pedoe \cite{tunstall91} also suggested a similar approach, without providing an implementation.

Gross \emph{et al.}~\cite{gross02} used a hybrid of genetic programming and evolution strategies to improve the efficiency of an already existing search algorithm using a distributed computing environment on the Internet.

In the following section, we present a novel approach that facilitates the use of GA for efficiently evolving the parameters of an evaluation function. As will be demonstrated, the method is very fast, and the evolved program is on par with today's strongest chess programs. 

\section{Expert-Driven Fitness Evaluation}

Due to the impediments already discussed, establishing fitness evaluation by means of playing numerous games is not practical. However, one can exploit a vast reservoir of previously under-utilized information. While the evaluation functions of existing chess programs are carefully-guarded secrets, it is standard practice for a chess program to (partially) reveal the score for any given position encountered in a game. We show in this section how to use genetic algorithms to essentially reverse engineer these evaluation functions. In particular, we show that such reverse engineering can be carried out very rapidly and successfully, and that a program based on an evaluation function learned from a particular expert, can perform as well as the expert. The program evolves its evaluation function by learning from an expert according to the steps shown in Figure \ref{fig:mentor}.

\begin{figure}[htbp]
\begin{center}
\line(1,0){320}

\begin{enumerate}
\item Generate a list of random problems.
\item For each problem, let the expert evaluate the problem and store the result.
\item Let each individual evaluate all the problems, and for each individual calculate the average difference (over all problems) between the value given by the individual and the value issued by the expert. The fitness of the individual will be inversely proportional to this average difference.
\end{enumerate}

\line(1,0){320}
\caption{Expert-driven fitness evaluation.}
\label{fig:mentor}
\end{center}
\end{figure}

In our case, each problem is associated with a chess position, and the expert input is the score of the evaluation function of a state-of-the-art chess engine. In other words, we generate a list of random chess positions for each generation, and let a strong chess engine evaluate all of them. Afterwards, we let the evaluation function of each of these individuals evaluate the positions. The closer the evaluation of an individual to the evaluation of the expert, the higher its fitness value. In the following subsections, we describe in detail the chess programs, the implementation of our expert-driven approach, and the GA parameters used.

\subsection{The Chess Programs}

We use \textsc{Falcon} chess engine as the ``expert'' for our experiments. \textsc{Falcon} is a 2700+ Elo rated grandmaster-level chess program, which has successfully participated in three World Computer Chess Championships. (See Appendix B for the Elo rating system.) \textsc{Falcon} uses \textsc{NegaScout}/PVS \cite{campbell83,reinfeld83} search, with null-move pruning \cite{beal89,david08b,donninger93}, internal iterative deepening \cite{anantharaman91,scott69}, dynamic move ordering (history + killer heuristic) \cite{akl77,gillogly72,schaeffer83,schaeffer89}, multi-cut pruning \cite{bjornsson98,bjornsson01}, selective extensions \cite{anantharaman91,beal95} (consisting of check, one-reply, mate-threat, recapture, and passed pawn extensions), transposition table \cite{nelson85,slate77}, futility pruning near leaf nodes \cite{heinz98a}, and blockage detection in endgames \cite{david06}. 

\textsc{Falcon}'s extensive evaluation function consists of more than 100 parameters, and its implementation contains several thousand lines of code. Our initial chromosomes, which are to evolve by mimicking the expert, use the exact same search techniques \textsc{Falcon} is using, and differ from \textsc{Falcon} only in their evaluation function, which consists of fewer than 40 parameters. In our experiments we randomly initialize the parameters of the organisms, thus resulting in a random evaluation function (i.e., no chess knowledge). The goal is to evolve these parameters by mimicking the behavior of the \textsc{Falcon}.

\subsection{Encoding the Evaluation Function}

Using \textsc{Falcon} as the expert, we evolve our organisms' evaluation function to mimic the behavior of the expert, thereby improving their strength. We use only the output of \textsc{Falcon}'s evaluation function, and otherwise make no assumption about the methods \textsc{Falcon} uses to compute this function. Thus, we only make use of \textsc{Falcon}'s scores to optimize the parameters of the organisms, not the parameter values of \textsc{Falcon}'s evaluation function, which (for our purposes) are considered unknown.

Although, as described above, our organisms' evaluation function consists of a much smaller number of parameters than \textsc{Falcon}'s, it does cover all important aspects of a position, e.g., material, piece mobility and centricity, pawn structure, and king safety. Despite this considerably simpler evaluation function, it can achieve comparable performance to \textsc{Falcon}'s, as shown in Section 4.

In order to demonstrate the effectiveness of our expert-driven approach, we ignore entirely the initial values of the evaluation function's parameters, and instead, assign random values to all of them. In other words, the initial organisms play like a novice with no knowledge about the game (other than the legal moves and certain built-in tactics). 

The parameters of the organisms' evaluation function are represented as a binary bit-string (chromosome size: 230 bits), initialized randomly. We further impose the restriction that except for the five parameters representing the material values of the pieces, all the other parameters are assigned a fixed length of 6 bits per parameter. Obviously, there are many parameters for which 3 or 4 bits suffice. However, allocating a fixed length of 6 bits to all parameters ensures that \emph{a priori} knowledge does not bias the algorithm in any way.

\subsection{Expert-Driven Fitness Function}

As already described, our goal is to evolve the parameters so that the evaluation function would produce as close a score as possible to \textsc{Falcon}'s evaluation function, given the same position. 

For our experiments, we use a database of 10,000 games by grandmasters of rating above 2600 Elo, and randomly pick one position from each game. Of these 10,000 positions, we select 5,000 positions for training and 5,000 for testing.

At first, we let \textsc{Falcon} search each of the 10,000 positions to a depth of 2 plies, and store its evaluation of the position. (Denote the expert's score for position $p$ by $S_{e,p}$.) Then, at each generation we randomly select 1,000 positions out of the 5,000 designated positions for the learning phase. This random selection of positions introduces additional variety in the test sets, which should help prevent premature convergence to suboptimal values.

For each organism we translate its chromosome bit-string into a corresponding evaluation function, and apply the evaluation function to each of the $N$ positions examined (in our case, $N=1000$). Let $S_{i,p}$ denote the score of organism $i$ for position $p$. For each position $p$ define the organism's error as

\begin{displaymath}
E_{i,p} = |S_{e,p} - S_{i,p}|,
\end{displaymath}

\noindent so the average overall error (for the organism) over the $N$ positions is given by

\begin{displaymath}
E_i = \frac{\displaystyle \sum_{p=1}^{N} E_{i,p}}{N}.
\end{displaymath}

Finally, the fitness value of organism $i$ is $F_i = -E_i$, i.e., the smaller the average error, the higher the fitness value.

\subsection{GA Parameters}

Other than the special fitness function described above, we use a standard implementation of GA with proportional selection and single point crossover. The following parameters are used:
\\
\\
population size = 1000\\
crossover rate = 0.75\\
mutation rate = 0.002\\
number of generations = 300\\

At the end of each generation, we replicate the best organism and delete the worst organism. Note that each organism is in fact a unique encoding of the evaluation function values.

In the following section we provide our experimental results, both in terms of the learning efficiency and the performance gain of the best evolved individual.

\section{Experimental Results}

We first present the results of running the expert-driven GA as described in the previous section. Then, we provide the results of several experiments that measure the strength of the evolved program in comparison to its original version.

\subsection{Learning Results}

\begin{figure}[htbp]
\centering
\epsfig{file=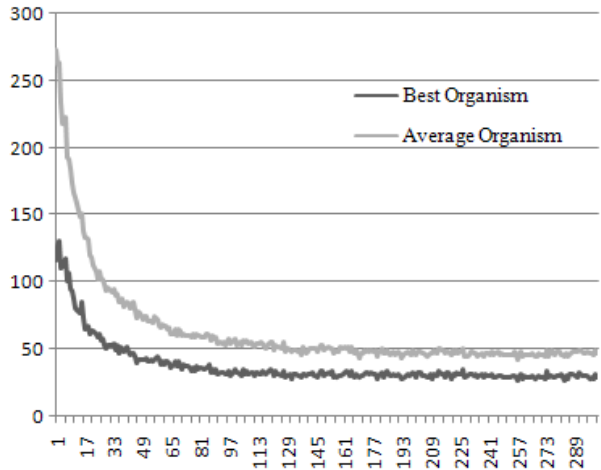, height=2.5in, width=3.1in}
\caption{Average error per position (in centipawns) for the best organism and the population average at each generation (total time for 300 generations: 442 seconds).}
\label{fig:graph}
\end{figure}

Figure~\ref{fig:graph} shows the average error per position of the best organism and the population average for 300 generations\footnote{An evaluation unit in chess programs is commonly called a \emph{centipawn}, i.e., 1/100th of the value of a pawn. Traditionally, a pawn is assigned a value of 100, and all other parameters are assigned relative values. However, the value of a pawn itself need not be exactly 100, so a unit of evaluation may no longer be exactly 1/100th of a pawn. Despite this inconsistency, the term centipawn is still used to denote the smallest evaluation unit.}. Specifically, the results indicate that the average error and the error of the best organism in the first few generations are greater than 250 centipawns and 130 centipawns, respectively. These large initial errors, that are due to the random parameter initialization, lead in the first few generations to very small fitness values for many organisms, and subsequently, to their rapid extinction.

Close to generation 35, the average error of the best organism drops below 50 centipawns. At this stage, large parameter values (such as piece material, etc.) are already well tuned for most of the organisms, and the smaller parameter values are fine tuned during the remaining generations. At generation 300, the average error of the best organism is 28 centipawns, and the average error in the population is 47 centipawns. Figure~\ref{fig:best-evolved} provides the evolved values of the best individual.

\begin{figure}[htbp]
\begin{center}
\begin{verbatim}
     PAWN_VALUE                          83
     KNIGHT_VALUE                       322
     BISHOP_VALUE                       323
     ROOK_VALUE                         478
     QUEEN_VALUE                        954
     PAWN_ADVANCE_A                       2
     PAWN_ADVANCE_B                       4
     PASSED_PAWN_MULT                     5
     DOUBLED_PAWN_PENALTY                21
     ISOLATED_PAWN_PENALTY               10
     BACKWARD_PAWN_PENALTY                3
     WEAK_SQUARE_PENALTY                  7
     PASSED_PAWN_ENEMY_KING_DIST          5
     KNIGHT_SQ_MULT                       7
     KNIGHT_OUTPOST_MULT                  8
     BISHOP_MOBILITY                      5
     BISHOP_PAIR                         44
     ROOK_ATTACK_KING_FILE               30
     ROOK_ATTACK_KING_ADJ_FILE            1
     ROOK_ATTACK_KING_ADJ_FILE_ABGH      21
     ROOK_7TH_RANK                       32
     ROOK_CONNECTED                       2
     ROOK_MOBILITY                        2
     ROOK_BEHIND_PASSED_PAWN             48
     ROOK_OPEN_FILE                      12
     ROOK_SEMI_OPEN_FILE                  6
     ROOK_ATCK_WEAK_PAWN_OPEN_COLUMN      7
     ROOK_COLUMN_MULT                     3
     QUEEN_MOBILITY                       0
     KING_NO_FRIENDLY_PAWN               27
     KING_NO_FRIENDLY_PAWN_ADJ           17
     KING_FRIENDLY_PAWN_ADVANCED1        12
     KING_NO_ENEMY_PAWN                  11
     KING_NO_ENEMY_PAWN_ADJ               3
     KING_PRESSURE_MULT                   8
\end{verbatim}

\caption{Evolved parameters of the best individual.}
\label{fig:best-evolved}
\end{center}
\end{figure}

With the completion of the learning phase, we used the additional 5,000 positions set aside for testing. We let the best evolved organism evaluate these positions, and compared its evaluation with that of the expert (\textsc{Falcon}). The average error in this case is 30 centipawns. This indicates that the first 5,000 positions used for training cover most types of positions that can arise, as the average error is very similar to the average error for the testing set. The entire 300-generation evolution lasted 442 seconds on our machine (see Appendix A), that is, less than 8 minutes.

The results clearly demonstrate that within a few minutes our GA-based module evolved from scratch an evaluation function whose parameters yield very similar performance to that of the expert.

\subsection{Performance of the Evolved Organism against the Expert}

We now provide the results of a series of matches between the programs. In order to obtain a baseline, we first observed the performance of a randomly initialized organism (which we call \textsc{RandOrg}) against the expert, \textsc{Falcon}, and the best evolved organism (which we call \textsc{Evol*}). We then conducted a series of games between \textsc{Falcon} and \textsc{Evol*}. Table~\ref{tab:results} provides the results. The matches \textsc{Falcon} vs.~\textsc{RandOrg} each and \textsc{Evol*} vs.~\textsc{RandOrg} consisted of 300 games played under a time limit of 3 minutes per game, and the match between \textsc{Evol*} and \textsc{Falcon} consisted of 1000 games played under a time limit of 10 minutes per game (i.e., a more extensive set of games and a longer time limit were used in order to obtain a more accurate assessment).

\begin{table}[htbp]

\begin{center}
\begin{tabular}{|l||c|c|c|}
\hline
Match & Result & W\% & RD\\
\hline
\hline
\textsc{Falcon} - \textsc{RandOrg} & 297.0 - 3.0 & 99.0\% & +798\\
\hline
\textsc{Evol*} - \textsc{RandOrg} & 296.0 - 4.0 & 98.7\% & +748\\
\hline
\textsc{Evol*} - \textsc{Falcon} & 544.5 - 455.5 & 54.5\% & $+31$\\
\hline
\end{tabular}
\end{center}
\caption{Results of the games between the three programs (W\% is the winning percentage, and RD is the Elo rating difference (see Appendix B)). Win = 1 point, draw = 0.5 point, and loss = 0 point.}
\label{tab:results}

\end{table}

The results of \textsc{Falcon} vs.~\textsc{RandOrg} show that the randomly initialized organism loses almost all the games, which is the expected outcome. Moreover, the evolved \textsc{Evol*} too resoundingly outperforms the randomly initialized organism\footnote{Note that the two programs (including the sets of parameters of their evaluation function) are essentially the same, except for the actual values assigned to these parameters.}, clearly demonstrating the immense improvement due to evolution.

The results further indicate that the evolved \textsc{Evol*} performes on par with the expert, \textsc{Falcon}. In particular, the results establish empirically that despite using an evaluation function with a smaller number of parameters, our expert-driven module, \textsc{Evol*}, evolves parameter values that yield comparable performance to \textsc{Falcon}'s. In fact, we cannot help but observe the curious fact that \textsc{Evol*}'s performance is actually a bit \emph{stronger} than \textsc{Falcon}'s. Indeed, at 95\% statistical confidence (2 standard deviations), the rating difference is 31$\pm$17 Elo, and at 99.7\% statistical confidence (3 standard deviations) the rating difference is 31$\pm$26 Elo. That is, the evolved \textsc{Evol*} is actually slightly superior to \textsc{Falcon} with a statistical confidence of over 99.7\% (see Appendix B for a detailed derivation). Apparently, the improved performance of the evolved organism over the expert can be attributed to the following (domain-specific) factors: (1) The evolved program's evaluation function has fewer parameters than the expert, which makes it capable of applying the evaluation function faster, thus resulting in a higher processing rate (i.e., searching more positions per second), and (2) when the program is evolved to mimic the behavior of the expert at a 2-ply search, its evaluation function is evolved to statically incorporate some of the dynamic knowledge of the expert.

\subsection{Performance of the Evolved Organism Against Other Programs}

We ran two additional series, each consisting of 300 games against the chess program \textsc{Crafty} (of Robert Hyatt \cite{hyatt90}). \textsc{Crafty} has successfully participated in numerous World Computer Chess Championships (WCCC), and is a direct descendent of \textsc{Cray Blitz}, the WCCC winner of 1983 and 1986. It is frequently used in the literature as a standard reference. Thus, we compared our evolved \textsc{Evol*}, and the expert, \textsc{Falcon}, against \textsc{Crafty}. Table~\ref{tab:crafty} provides the results.

\begin{table}[htbp]

\begin{center}
\begin{tabular}{|l||c|c|c|}
\hline
Match & Result & W\% & RD\\
\hline
\hline
\textsc{Falcon} - \textsc{Crafty} & 173.5 - 126.5 & 57.8\% & +55\\
\hline
\textsc{Evol*} - \textsc{Crafty} & 177.0 - 123.0 & 59.0\% & +63\\
\hline
\end{tabular}
\end{center}
\caption{\textsc{Crafty} vs.~\textsc{Evol*} and \textsc{Falcon} (W\% is the winning percentage, and RD is the Elo rating difference).}
\label{tab:crafty}

\end{table}

The results show that the evolved \textsc{Evol*} is clearly superior to \textsc{Crafty}. Also, the relative performance of  \textsc{Falcon} and \textsc{Evol*} against \textsc{Crafty}, implies again that \textsc{Evol*} is slightly stronger than \textsc{Falcon}.

This phenomenon was observed in yet another experiment. For measuring the tactical strength of the programs, we used the Encyclopedia of Chess Middlegames (ECM) test suite, consisting of 879 positions. Each program was given 5 seconds per position to come up with the ``correct'' move for the position. Table~\ref{tab:ecm} provides the results. As can be seen, \textsc{Evol*} solved significantly more problems than \textsc{Crafty} and a few more than \textsc{Falcon}. 

\begin{table}[htbp]
\begin{center}
\begin{tabular}{|c|c|c|}
\hline
\textsc{Evol*} & \textsc{Falcon} & \textsc{Crafty}\\
\hline
\hline
649 & 645 & 593\\
\hline
\end{tabular}
\end{center}
\caption{Number of ECM positions solved by each program (time: 5 seconds per position).}
\label{tab:ecm}
\end{table}

Finally, we extended our experiments to compare the performance of \textsc{Evol*} against several of the world's top commercial chess programs. These programs included \textsc{Junior}, \textsc{Fritz}, and \textsc{Hiarcs}. \textsc{Junior} won the World Microcomputer Chess Championship in 1997 and 2001 and the World Computer Chess Championship in 2002, 2004, and 2006. In 2003 \textsc{Junior} played a 6-game match against former world champion Garry Kasparov that resulted in a 3--3 tie. In 2007 \textsc{Junior} won the ``ultimate computer chess challenge'' organized by the World Chess Federation (FIDE), defeating \textsc{Fritz} 4--2. In 1995 \textsc{Fritz} won the World Computer Chess Championship. In 2002, \textsc{Fritz} drew the ``Brains in Bahrain'' match against the former world champion Vladimir Kramnik 4?-4, in 2003 it drew a four-game match against Garry Kasparov, and in 2006 it defeated Vladimir Kramnik 4--2. \textsc{Hiarcs} won the 1993 World Microcomputer Chess Championship, and in 2003 played a four-game match against Grandmaster Evgeny Bareev, then the 8th ranked player in the world. All the four games ended in a draw, resulting in a tied match. In 2007 \textsc{Hiarcs} won the 17th International Paderborn Computer Chess Championship.

Table~\ref{tab:commercial} provides the results against these top commercial programs. Note that \textsc{Evol*} was evolved by learning once from \textsc{Falcon} (and not from the program it played against).

\begin{table}[htbp]

\begin{center}
\begin{tabular}{|l||c|c|c|}
\hline
Match & Result & W\% & RD\\
\hline
\hline
\textsc{Evol*} - \textsc{Junior} & 135.0 - 165.0 & 45.0\% & $-35$\\
\hline
\textsc{Evol*} - \textsc{Fritz} & 154.0 - 146.0 & 51.3\% & +9\\
\hline
\textsc{Evol*} - \textsc{Hiarcs} & 172.5 - 127.5 & 57.5\% & +52\\
\hline
\end{tabular}
\end{center}
\caption{\textsc{Evol*} vs.~\textsc{Junior}, \textsc{Fritz}, and \textsc{Hiarcs} (W\% is the winning percentage, and RD is the Elo rating difference).}
\label{tab:commercial}

\end{table}

The results show that the performance of genetically evolved program is on par with that of the top commercial chess programs, outperforming \textsc{Hiarcs} by 52 Elo points, obtaining an almost equal score against \textsc{Fritz}, and being slightly outperformed by \textsc{Junior} (by 32 Elo points).

In addition, Table~\ref{tab:ecm2} compares the tactical performance of our evolved organism against these three commercial programs. The results show the number of ECM positions solved by each program. A similar trend emerges, i.e., the evolved organism is on par with \textsc{Fritz} and \textsc{Hiarcs} in terms of the tactical strength, and slightly inferior to \textsc{Junior}.

\begin{table}[htbp]
\begin{center}
\begin{tabular}{|c|c|c|c|}
\hline
\textsc{Evol*} & \textsc{Junior} & \textsc{Fritz} & \textsc{Hiarcs}\\
\hline
\hline
649 & 681 & 640 & 642\\
\hline
\end{tabular}
\end{center}
\caption{Number of ECM positions solved by each program (time: 5 seconds per position).}
\label{tab:ecm2}
\end{table}

Note that all the experiments described above were conducted on a \emph{uniform} platform, i.e., for each match both programs ran on the same machine, and were allocated the same resources (e.g., same memory size, opening book, endgame tablebases, etc.). In the next subsection we report on the performance of our evolved organism in a recent World Computer Chess Championship, which was not conducted on a uniform platform.

\subsection{Performance in the 2008 World Computer Chess Championship}

Using our expert-driven approach, we participated with a genetically evolved version of our program in the 2008 World Computer Chess Championship in Beijing, China. Competing with an average laptop against 9 of the strongest programs in the world (8 of which ran on fast multicore machines ranging from 4 to 40 cores), our program reached 2nd place in the World Computer Speed Chess Championship and 6th place in the World Computer Chess Championship. These highly surprising results, especially in light of the huge hardware handicap, in comparison to our competitors, demonstrate the capabilities of our expert-driven approach.

Table~\ref{tab:wccc} provides the list of competitors, the number of processors/cores utilized, and the result of our genetically evolved program against each competitor.

\begin{table}[htbp]
\begin{center}
\begin{tabular}{|l||c|c|c|}
\hline
Program & Number of Cores & WCCC Result & WCSCC Result\\
\hline
\hline
\textsc{Rybka} & 40 & $-$ & $+$\\
\hline
\textsc{Cluster Toga} & 24 & $+$ & $+$\\
\hline
\textsc{Jonny} & 16 & $=$ & $+$\\
\hline
\textsc{Junior} & 12 & $=$ & $=$\\
\hline
\textsc{Hiarcs} & 8 & $-$ & $-$\\
\hline
\textsc{Shredder} & 8 & $-$ & $=$\\
\hline
\textsc{The Baron} & 4 & $+$ & $=$\\
\hline
\textsc{Sjeng} & 4 & $-$ & $=$\\
\hline
\textsc{Mobile Chess} & 1 & $+$ & $+$\\
\hline
\end{tabular}
\end{center}
\caption{Results of our genetically evolved program (using one core) against each of the competitors in the 2008 World Computer Speed Chess Championship (WCSCC) and World Computer Chess Championship (WCCC); ``+'' stands for a victory for our program, ``$-$'' stands for a loss, and ``='' stands for a draw.}
\label{tab:wccc}
\end{table}

The results in Table~\ref{tab:wccc} show that our evolved organism managed to defeat several programs running on markedly faster machines (up to 40 times the speed of our platform).

\section{Concluding Remarks and Future Research}

In this paper, we presented a novel expert-driven approach for efficient automatic tuning of the parameters of a chess program's evaluation function. Wherever an intelligent entity already exists, we can employ it as an expert within our GA-based framework to evolve organisms that mimic its behavior. In other words, our approach enables duplicating the behavior of another intelligent organism by observing merely its performance, without accessing its underlying mechanism.

According to our experiments, organisms evolved within a few minutes from randomly initialized chromosomes to sets of highly-tuned parameters that yield similar performance to that of the expert, with respect to the same set of positions. The results of the games played demonstrate the significant gain of the evolved version, which clearly outperforms its original version. Note that the successful duplication of the expert's behavior was achieved despite the fact that the evaluation function of the evolved program consists of a considerably smaller number of parameters.

In this extended version of our previously presented work \cite{david08a}, we included an  extended set of experiments to assess more accurately the performance of the evolved program. Specifically, we measured the performance gain due to evolution by comparing a random organism against the evolved organism and the the expert, ran a longer series of matches between the evolved organism and the expert, compared the performance of the evolved organism against three top commercial chess programs, and observed the tactical performance of the evolved organism against several top programs. Finally, we provided a detailed account of our participation in the World Computer Chess Championship, where despite a huge hardware disadvantage, our genetically evolved program achieved second place in a recent World Computer Speed Chess Championship, and sixth place in the World Computer Chess Championship. These extended results firmly establish the merit of our GA-based method for automatically learning the parameters of a chess program's evaluation function.

For future research, we intend to develop additional capabilities based on the presented expert-driven approach. In this paper we focused on how another computer program can serve as an expert. However, using human players as experts is a more difficult challenge, as there is no explicit notion of a numerical evaluation of a position. We believe, though, that a record of hundreds of games of a human player would provide sufficient data for similar learning to take place. One method we intend to explore, is to extract several thousand positions from games played by a human expert, and for each position assign higher fitness for the organism that produces the move played by the expert. If successful, this approach would basically enable the program to perform like the expert, without ``probing'' his/her ``mind''. For example, we might be able to develop a program that plays like Kasparov just by learning from his games.

In this work we used a single expert. An alternative implementation might employ several experts, using the ``wisdom of crowds'' concept to evolve an individual which is ``wiser'' than the experts. It is well known that each chess program has its strengths and weaknesses. By employing several expert chess engines, it might be possible to combine the strengths of all of them, and outperform each individual expert.

Our expert-driven approach could also be applied to the problem of player recognition. Given a set of $N$ players, the simplest approach is to separately evolve $N$ organisms, each mimicking the behavior of one of the players, respectively. Then, given a query game (played by one of the $N$ players), we would let each of the generated organisms evaluate the position. The player whose cloned organism agrees most closely with the moves made, is most likely to have played the game in question.

Finally, we believe that the approach pursued in this paper for parameter tuning could be applied to a wide array of problems in which the output of an expert's evaluation function is available for training purposes.

\section*{Appendix}
\appendix

\section{Experimental Setup}

\hspace*{-4pt}Our experimental setup consisted of the following resources:

\begin{itemize}
\item \textsc{Falcon} chess engine running under UCI protocol, and \textsc{Crafty 19}, \textsc{Junior 9}, \textsc{Fritz 8}, and \textsc{Hiarcs 8} running as a native ChessBase engines.

\item Encyclopedia of Chess Middlegames (ECM) test suite, consisting of 879 positions.

\item Fritz 8 interface for automatic running of matches. Fritz opening book was used for all games.

\item AMD Athlon 64 3200+ with 1 GB RAM and Windows XP operating system.
\end{itemize}

\section{Elo Rating System}

The Elo rating system, developed by Arpad Elo, is the official system for calculating the relative skill levels of players in chess. The following statistics from the January 2009 FIDE rating list provide a general impression of the meaning of the Elo rating system:

\begin{itemize}
\item 21079 players have a rating above 2200 Elo.
\
\item 2886 players have a rating between 2400 and 2499, most of whom have either the title of International Master (IM) or Grandmaster (GM).
\
\item 876 players have a rating between 2500 and 2599, most of whom have the title of GM.
\
\item 188 players have a rating between 2600 and 2699, all of whom have the title of GM.
\
\item 32 players have a rating above 2700.
\end{itemize}

Only four players have ever had a rating of 2800 or above. A novice player is generally associated with rating values below 1400 Elo. Given the rating difference (RD) between player A and player B, the expected winning rate $w$ ($0 \le w \le 1$) of player A is given by

\begin{equation}
w = \frac{1}{10^{-RD/400} + 1}.
\end{equation}

Given the winning rate of player A against player B (as is the case in our experiments), the expected rating difference between the two players can be derived from the above formula, i.e.,

\begin{equation}
RD = -400 \log_{10}(\frac{1}{w} - 1).
\end{equation}

In addition, given the results of a series of $N$ matches between two players, we can derive confidence intervals for their rating difference. Without loss of generality, let $W$, $D$, and $L$ denote, respectively, the number of wins, draws, and losses of the first player. The mean score and standard deviation are given, respectively, by

\begin{equation}
\overline{x} = \frac{W + D/2}{N}.
\end{equation}

and

\begin{equation}
s = \sqrt{\frac{W \cdot (1 - \overline{x})^2 + D \cdot(0.5 - \overline{x})^2 + L \cdot \overline{x}^2}{N - 1}}.
\end{equation}

Note that $\overline{x}$ is essentially an estimate of the expected winning rate. Now, suppose that we are interested in computing, for example, the 95\% confidence interval (which corresponds to $\pm$ two standard deviations) of the rating difference. For this we compute the lower and upper ends of the winning rate, i.e., $w_{lo} = \overline{x} - 2s$ and $w_{hi} = \overline{x} + 2s$. Substituting $w_{lo}$ and $w_{hi}$ in Eq.~(B.2) we obtain the corresponding lower and upper ends of the 95\% confidence interval of the rating difference. Given any confidence level, one can computer the corresponding RD confidence interval similarly to the above described steps.

\end{document}